\documentclass[conference,a4paper]{IEEEtran}
\usepackage{geometry}
\usepackage{cite}
\usepackage[pdftex]{graphicx}
\usepackage{amsmath}

\usepackage{algorithm} % No
\usepackage{algorithmicx} % Si
\usepackage[noend]{algpseudocode}
\usepackage{adjustbox}

\usepackage[percent]{overpic}
\usepackage{xspace}
\usepackage[table]{xcolor}
\usepackage{rotating}
\usepackage{sidecap}

\newcommand{\comment}[1]{}
\newcommand{\dt}{{Saddle}\xspace}
\algnewcommand\algorithmicinput{\textbf{Input:}}
\algnewcommand\INPUT{\item[\algorithmicinput]}
\algnewcommand\algorithmicoutput{\textbf{Output:}}
\algnewcommand\OUTPUT{\item[\algorithmicoutput]}

\newcommand{\p}{\ensuremath{\mathbf{p}}\xspace}

\makeatletter
\DeclareRobustCommand\onedot{\futurelet\@let@token\@onedot}
\def\@onedot{\ifx\@let@token.\else.\null\fi\xspace}

\def\eg{\emph{e.g}\onedot}

\def\etal{\emph{et al}\onedot}

\def\ccaEF#1{\cellcolor{black!#1!black!#1}\ifnum #1>50\color{white}\fi{#1}}
\def\ccaEVD#1{\cellcolor{black!#10}\ifnum #1>15\color{white}\fi{#1}}
\def\ccalostinpast#1{\cellcolor{black!#1}\ifnum #1>70\color{white}\fi{#1}}
\def\ccaoxford#1{\cellcolor{black!#1!black!#1}\ifnum #1>50\color{white}\fi{#1}}
\def\ccasnavely#1{\cellcolor{black!#1!black!#1}\ifnum #1>50\color{white}\fi{#1}}
\def\ccasymbench#1{\cellcolor{black!#1!black!#1}\ifnum #1>50\color{white}\fi{#1}}
\def\ccastewart#1{\cellcolor{black!#1}\ifnum #1>55\color{white}\fi{#1}}
\makeatother

\newcolumntype{R}[2]{%
    >{\adjustbox{angle=#1,lap=\width-(#2)}\bgroup}%
    l%
    <{\egroup}%
}

\begin{document}

\newcommand*\rot{\multicolumn{1}{R{90}{1em}}}% no optional argument here, please!

\title{In the Saddle: Chasing Fast and Repeatable Features}

\newcommand{\hs}{\hspace{25pt}}
\author{\IEEEauthorblockN{Javier Aldana-Iuit, Dmytro Mishkin, Ond\v{r}ej Chum and Ji\v{r}\'{\i} Matas}
\IEEEauthorblockA{Center for Machine Perception, Czech Technical University in Prague\\
Prague, Czech Republic. \\
Email: \{aldanjav, mishkdmy, chum, matas\}@cmp.felk.cvut.cz}
}

\maketitle

\begin{abstract}

A novel similarity-covariant feature detector that 
extracts points whose neighborhoods, when treated as a 3D intensity surface, have a saddle-like intensity profile. The saddle condition is verified efficiently by intensity comparisons on two concentric rings that must have exactly two dark-to-bright and two bright-to-dark transitions satisfying certain geometric constraints.

Experiments show that the \dt features are general, evenly spread and appearing in high density in a range of images. 
The \dt detector is among the fastest proposed. 
In comparison with detector with similar speed, the \dt features show superior matching performance on number of challenging datasets.   
  
\end{abstract}

\IEEEpeerreviewmaketitle
% ------------------------------------------------------------------------
\section{Introduction}

Local invariant features\footnote{a.k.a. interest points, keypoints, feature points, distinguished regions.}  have  a wide range of applications:
image alignment and retrieval~\cite{Schmid1997,Philbin07}, specific object recognition~\cite{Lowe1999,Obdrzalek2005},
3D reconstruction~\cite{Frahm2010,Agarwal2011},
robot location~\cite{Se2002},
tracking by detection~\cite{ozuysal2006feature}, 
augmented reality~\cite{Lepetit2003}, etc. It is therefore not surprising that literally hundreds  of local feature detectors have been proposed~\cite{Mikolajczyk2004,Tuytelaars2008}.

In an application, the suitability of a particular local feature detector depends typically on more than
one property. The important characteristics are most commonly the {\it repeatability} -- the ability to respond to the same scene pre-image irrespective of changing acquisition conditions, {\it distinctiveness} -- the discriminative power of the intensity patches it extracts, {\it density} -- the number of responses per unit area, both average and maximum achievable and {\it efficiency} -- the speed with which the features are extracted. Other properties like the {\it  generality} of the scenes where the feature exhibits acceptable performance of the major characteristics, the evenness of the {\it coverage} of image, the  geometric {\it accuracy} are considered less often.

Local feature detectors with the largest impact lie on the "convex envelope" of the properties.
The Difference-of-Gaussians~\cite{Lowe1999} and the Hessian, either in the rotation~\cite{Beaudet1978}, similarity~\cite{Mikolajczyk2004} or affine covariant~\cite{Mikolajczyk2002} form, are arguably the most general detectors with high repeatability~\cite{Mikolajczyk2005}. For their efficiency, SURF~\cite{Bay2006}, FAST~\cite{Rosten2006} and ORB~\cite{Rublee2011} are the preferred choice for real-time applications or in cases when computational resources are limited as on mobile devices. MSERs~\cite{Matas2002} are popular for matching  of images with extreme viewpoint changes \cite{Mishkin2015MODS} and in some niches like text detection~\cite{Neumann2013,li2014characterness}. Learned detectors, trained to specific requirements like insensitivity to gross illumination changes, outperform ,in their domains, generic detectors~\cite{Verdie15}. For some problems, like matching between different modalities, any single detector is inferior to a combination of different local feature detectors~\cite{Mishkin2015WXBS}.

As a necessary condition of an interest point~\cite{Harris88}, the patch around the interest point must be dissimilar to patches in its immediate neighborhood. There are at least three types of such interest regions: (i) corners such as Harris corner detector~\cite{Harris88}, (ii) blobs such as MSER~\cite{Matas2002}, DoG~\cite{Lowe1999} or Hessian with positive determinant~\cite{Mikolajczyk2004}, and (iii) saddle points, \eg Hessian with negative determinant~\cite{Mikolajczyk2004}. Rapid detectors of corner points FAST~\cite{Rosten2006} and ORB~\cite{Rublee2011} and of blobs SURF~\cite{Bay2006} have been already proposed and are used in applications with significant time constraints.

\begin{figure}
\newcommand{\ic}[1]{\setlength{\fboxsep}{0pt}%
\parbox{.32\linewidth}{\centering \fbox{\includegraphics[width=0.99\linewidth]{images/samp_#1_loc}}\\\fbox{\includegraphics[width=0.99\linewidth]{images/samp_#1_tem}}\\\fbox{\includegraphics[width=0.99\linewidth]{images/samp_#1_3d}}}}
\ic{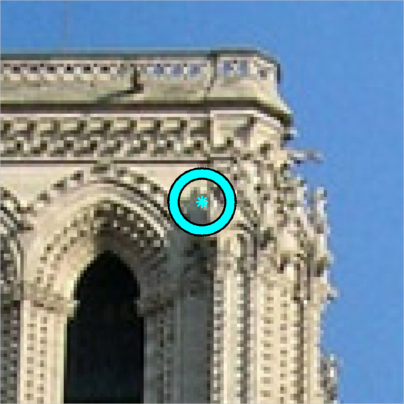}
\ic{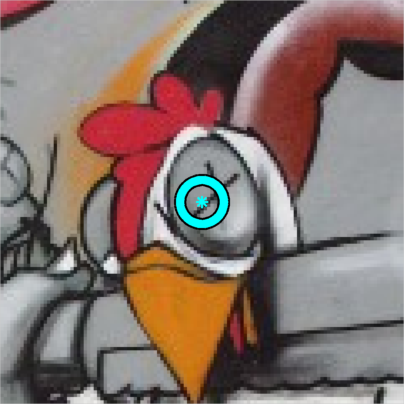}
\ic{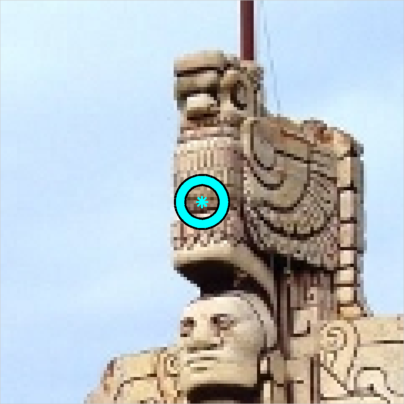}
\caption{\dt feature examples (top row). Corresponding image patches with accepted arrangements of dark (marked red), bright (green) and intermediate (blue) pixel intensities (middle row).  Pixel intensities around \dt points visualized as a 3D surface (bottom row).}
\label{fig.saddle}
\end{figure}

\begin{SCfigure}
\caption{The 8 pixel positions marked red form the \textit{inner ring} and the 16 positions $b_j$ marked blue form the \textit{outer ring}. Positions shared by both rings are bicolored.}

\centering
\setlength{\fboxsep}{0pt}%
\parbox{.42\linewidth}{\centering {\begin{overpic}[width=\linewidth]{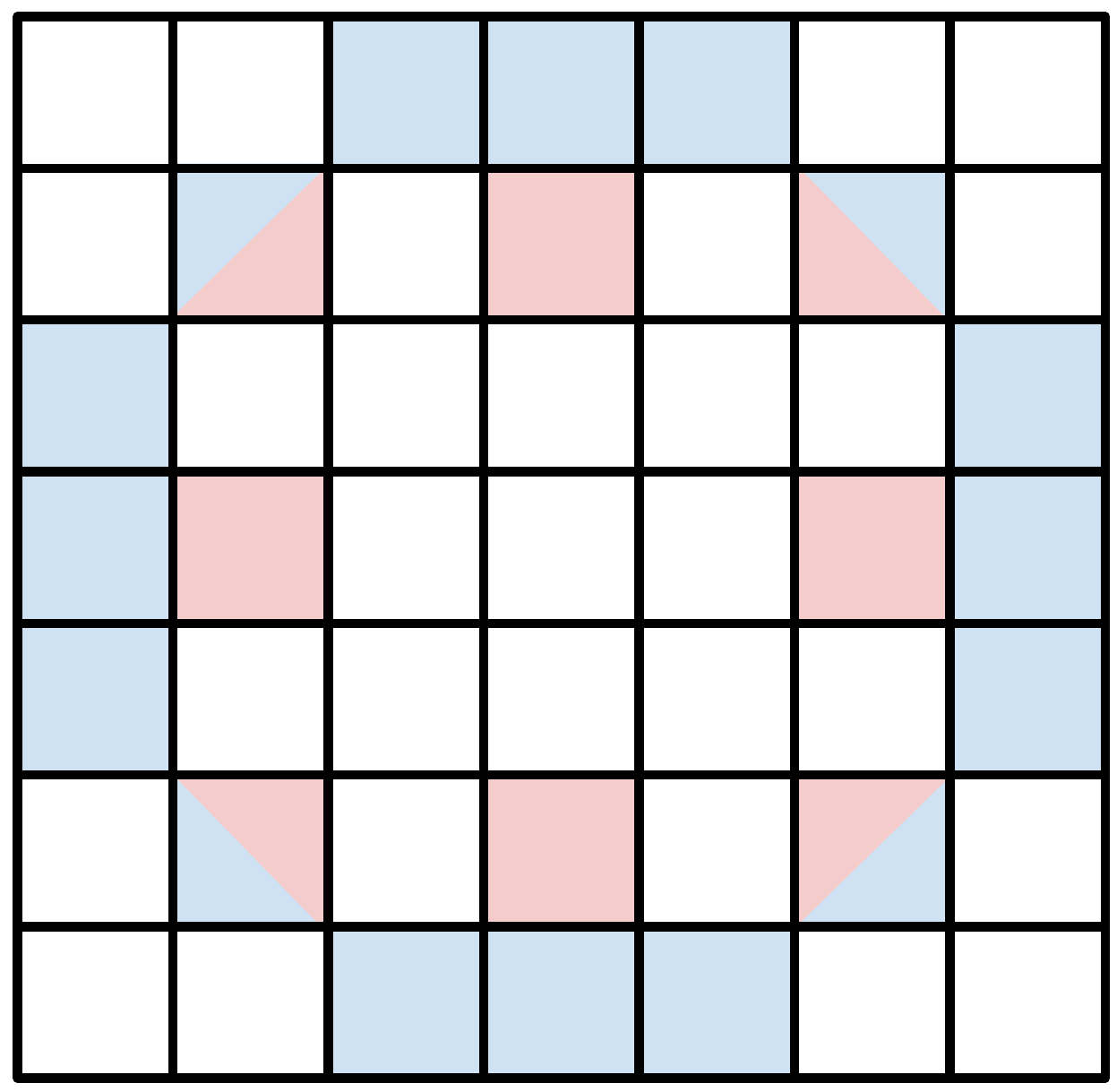}
% Center
\put (47,47) {\small$p$}
% Outer
\put (47,6) {\small$b_1$}
\put (60,6) {\small$b_2$}
\put (31,6) {\small$b_{16}$}

\put (74,19) {\small$b_3$}
\put (74,74) {\small$b_7$}
\put (18,74) {\small$b_{11}$}
\put (17,19) {\small$b_{15}$}

\put (88,34) {\small$b_4$}
\put (88,46) {\small$b_5$}
\put (88,60) {\small$b_6$}

\put (46,88) {\small$b_9$}
\put (60,88) {\small$b_8$}
\put (31,88) {\small$b_{10}$}

\put (3,34) {\small$b_{14}$}
\put (3,46) {\small$b_{13}$}
\put (3,60) {\small$b_{12}$}

\end{overpic}}}
\label{fig:innerouter}
\end{SCfigure}
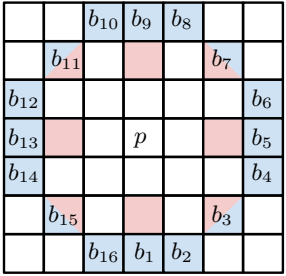

In this paper, we propose a novel similarity-covariant local feature detector called \dt.
The detector extracts points whose neighborhoods, when treated as a 3D intensity surface, have concave and convex profiles in a pair of orthogonal directions, see Fig.~\ref{fig.saddle}; in a continuous setting the points would  have a negative determinant of the Hessian matrix.  The saddle condition is verified on two concentric approximately circular rings which must have exactly two dark-to-bright and two bright-to-dark transitions satisfying certain geometric constraints, see Fig.~\ref{fig:innerouter}.

Experiments show that such points exist with high density in a broad class of images, are repeatably detectable, distinctive and are accurately localized. The \dt points are stable with respect to scale and thus a coarse pyramid is sufficient for their detection, saving time and memory. \dt is faster than SURF, a popular choice of detector when fast response is required, but slower than ORB.  Overall, the \dt detector provides an attractive combination of properties sufficient to have impact even in the mature area of local feature detectors.
 
\dt falls into the class of detectors that are defined in terms of intensity level comparisons, together with BRISK~\cite{Leutenegger2011}, FAST~\cite{Rosten2006}, its similarity-covariant extension ORB~\cite{Rublee2011},  and its  precursors like SUSAN~\cite{Smith1997} and the Trajkovic-Hedley detector~\cite{Trajković1998}. With the exception of BRISK, the intensity-comparison based detector aim at corner-like features and 
can be interpreted as a fast approximation of the Harris interest point detector~\cite{Harris88}\footnote{In fact, the ORB final interest point selection is a function of the Harris response computed on points that pass a preliminary test.}.
\dt is novel in that it uses intensity comparisons for detection of different local structures, related to Hessian rather than the Harris detector.
%%%%% Figure. Inner and outer template window
%%%%% Figure. Inner and outer template window
%%%%% Figure. Inner and outer template window
\begin{figure}
\newcommand{\ic}[1]{\parbox{.47\linewidth}{\centering \fbox{\includegraphics[width=0.99\linewidth]{images/tem_#1}}}}

\centering
\setlength{\fboxsep}{0pt}%
\parbox{.48\linewidth}{\centering \includegraphics[width=\linewidth]{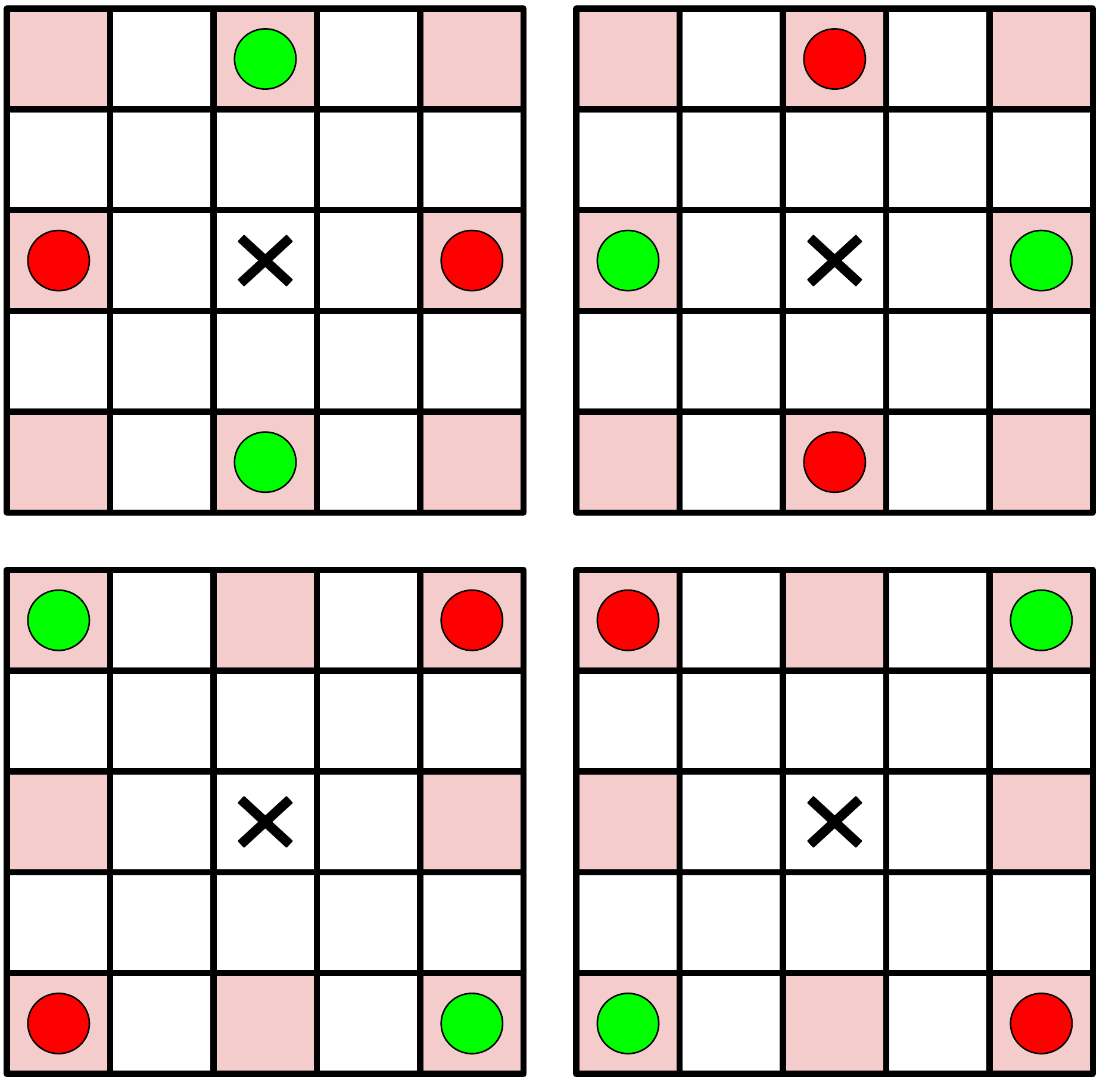}\\(a)}
\parbox{.48\linewidth}{\centering \ic{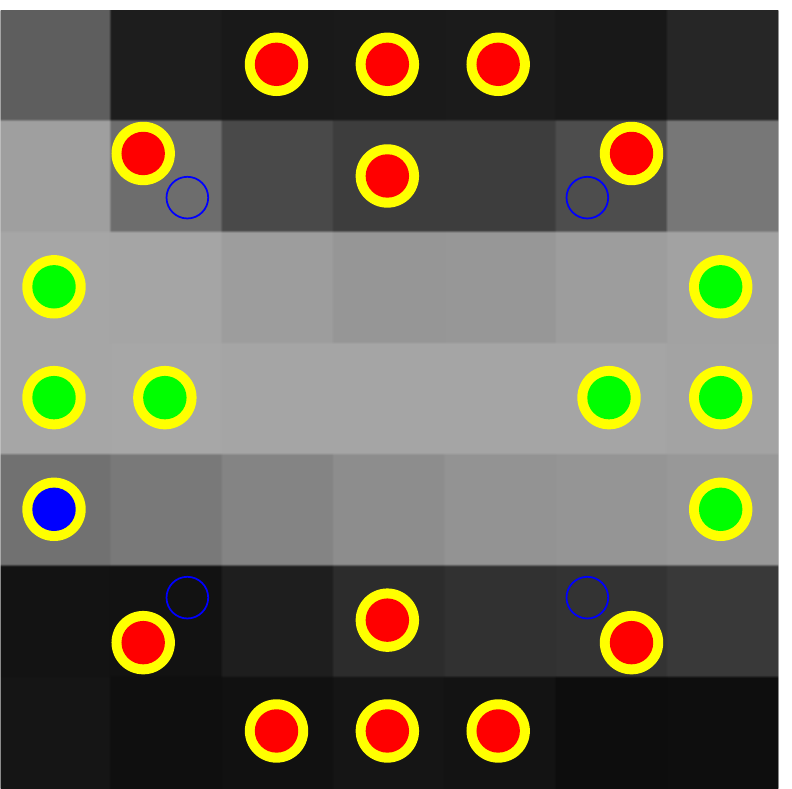} \ic{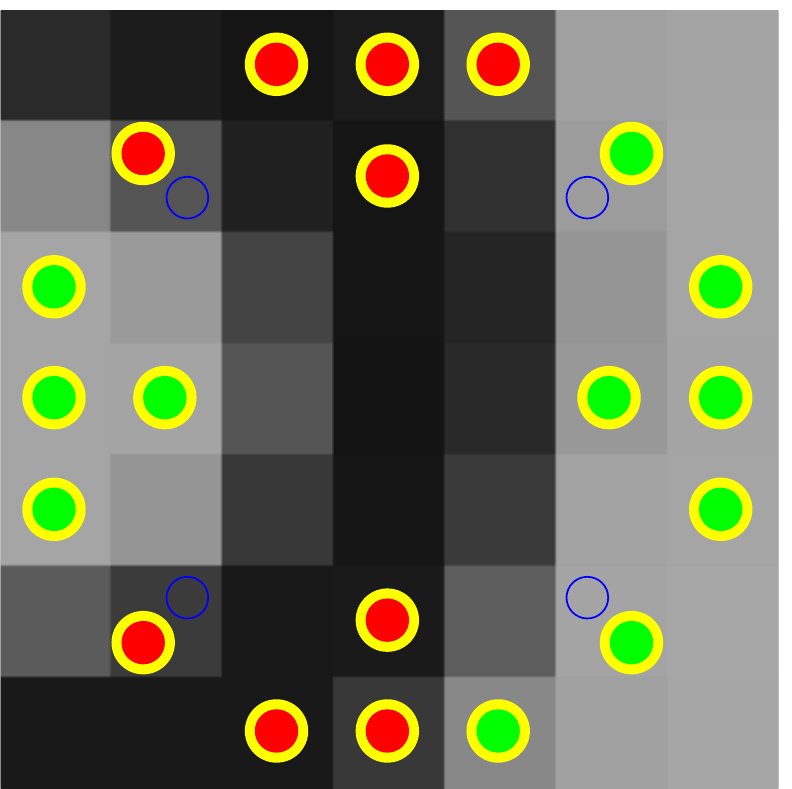}\\\vspace{0.1cm} \ic{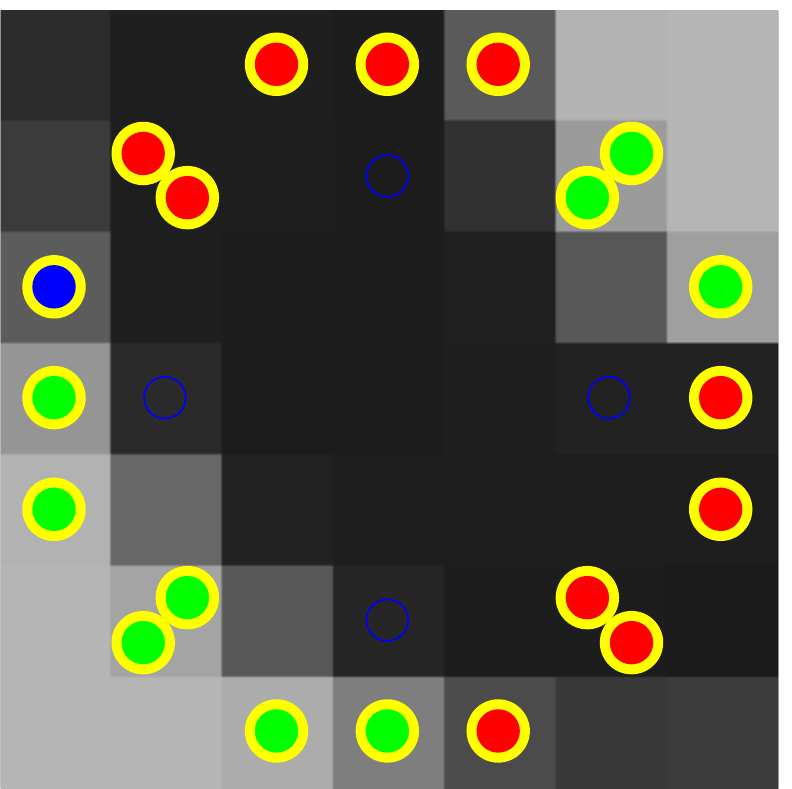} \ic{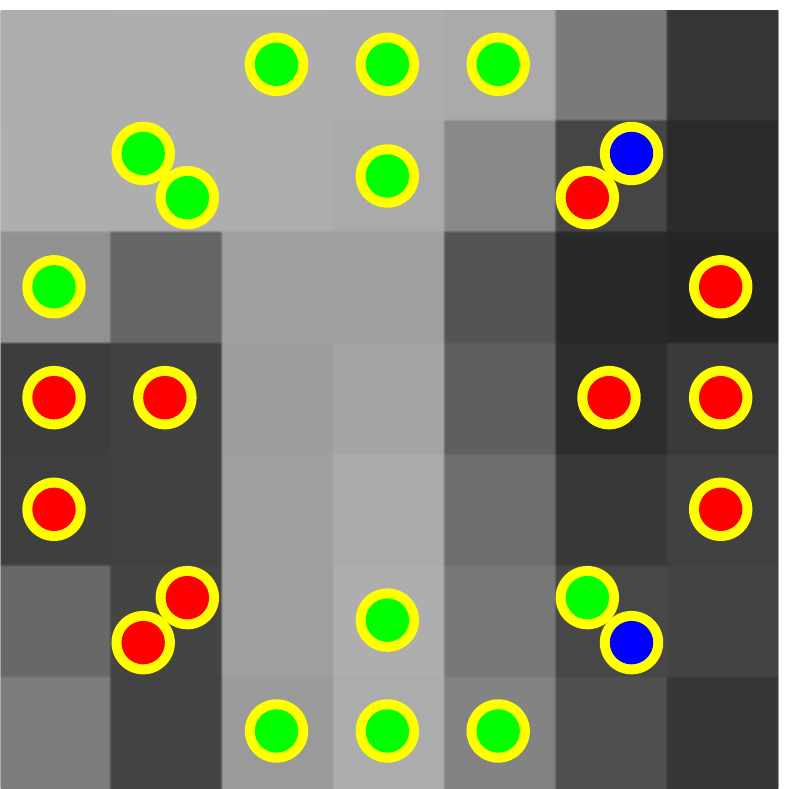}\\\vspace{0.2cm}(b) }

\caption{(a) The fast test for an alternating pattern on the inner ring. In each of the four patterns, green dots depict pixels with intensity strictly brighter than the intensity of pixels marked red. The location is eliminated further by \dt if none of the patterns is observed. (b) Examples of accepted patterns.}
\label{fig:innerring}
\end{figure}

% ------------------------------------------------------------------------

\section{The \dt Interest Point Detector}
The algorithmic structure of the  \dt keypoint detector is simple.
Convariance with  similarity transformation is achieved by localizing the keypoints in a scale-space pyramid~\cite{Lindeberg91a}. At every level of the pyramid, the \dt points are extracted in three steps. First, a fast alternating pattern test is performed in the inner ring, see Figs.~\ref{fig:innerouter} and \ref{fig:innerring}. This test eliminates about 80--85\% of the candidate points. If a point passes the first test, an alternating pattern test on the outer ring is carried out. Finally, points that pass both tests enter the post-processing stage, which includes non-maxima suppression and response strength selection. The algorithm is summarized in Alg.~\ref{alg.saddle_detection}.

\subsection {Alternating pattern on the inner ring}

%%%% SADDLE algorithm
% Algorithm overview
\begin{algorithm}[t]
\caption{Saddle feature detection}
\label{alg.detection}
\begin{algorithmic}
\INPUT Image $I$, $\epsilon$
\OUTPUT Set $F$ of \dt{} keypoints
\For {pyramid level $I_n$}
\For {every pixel $\p$ in $I_n$}
\If {$\sim$ INNER($\p$)}
\State continue
\EndIf
\State Compute $\rho_\p$
\If {$\sim$ OUTER($\p$, $\rho_\p$, $\epsilon$)}
\State continue
\EndIf
\State Compute response $R(\p)$
\EndFor
\State Non-Maxima Supression
\State Coordinate Refinement
\EndFor
\Return
\end{algorithmic}
\label{alg.saddle_detection}
\end{algorithm}

The first test is designed to be very fast and to reject majority of points. The test operates on pixels surrounding the central point -- the pink square in Fig~\ref{fig:innerouter}. In the test, two pairs of orthogonal directions are considered, one in the shape of $+$ and the other in the shape of $\times$. The test is passed if both points on the inner ring in one direction are strictly brighter than both points in the orthogonal direction. The four cases for passing the test are depicted in Fig.~\ref{fig:innerring}~(a). Note that either of the $+$ and $\times$ shapes can pass the test, or both.  

From the intensity values of the pixels satisfying the inner patter test, either four or eight pixels, depending whether one or both patterns passed the test, central intensity value $\rho$ is estimated. As a robust estimate, the median of the intensity values is used.

 %-------------------------------------------------------------------------

\begin{figure}
\centering
\newcommand{\ir}[2]{\setlength{\fboxsep}{0pt}%
\setlength{\fboxrule}{1pt}%
\parbox[c]{2\baselineskip}{\rotatebox{90}{#2}}\parbox[c]{.8\linewidth}{\fbox{\includegraphics[width=\linewidth]{images/#1}}}}
\ir{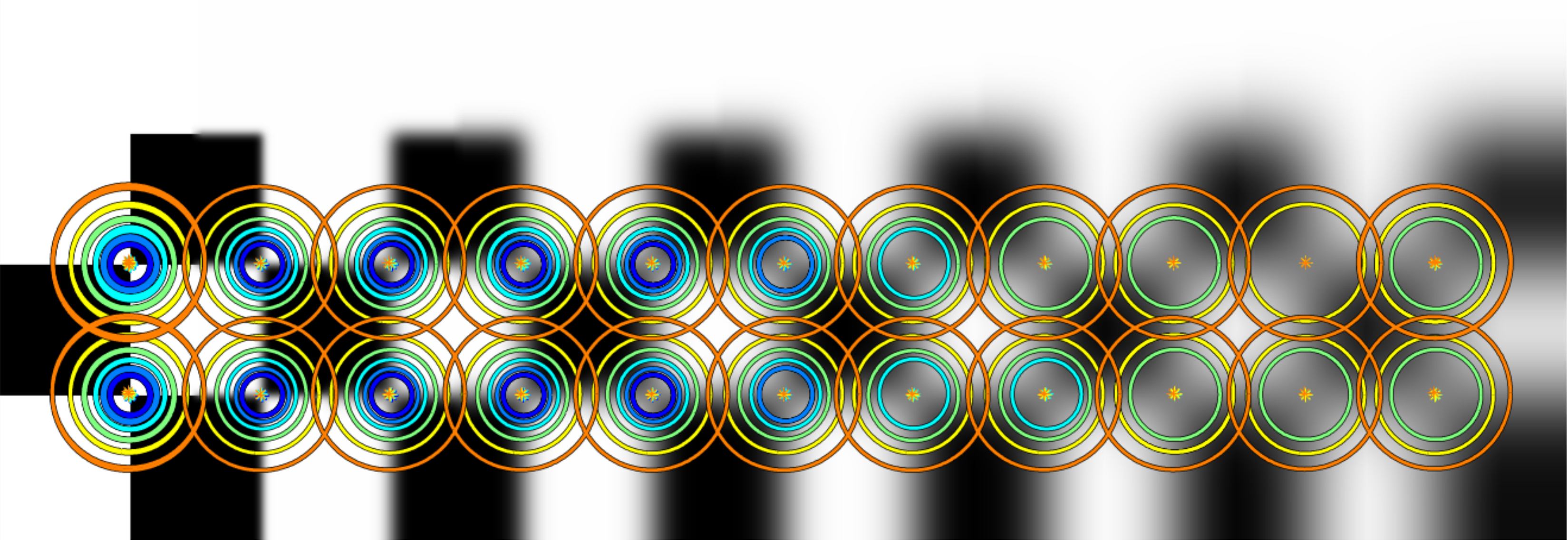}{\dt}\\[.1\baselineskip]
\ir{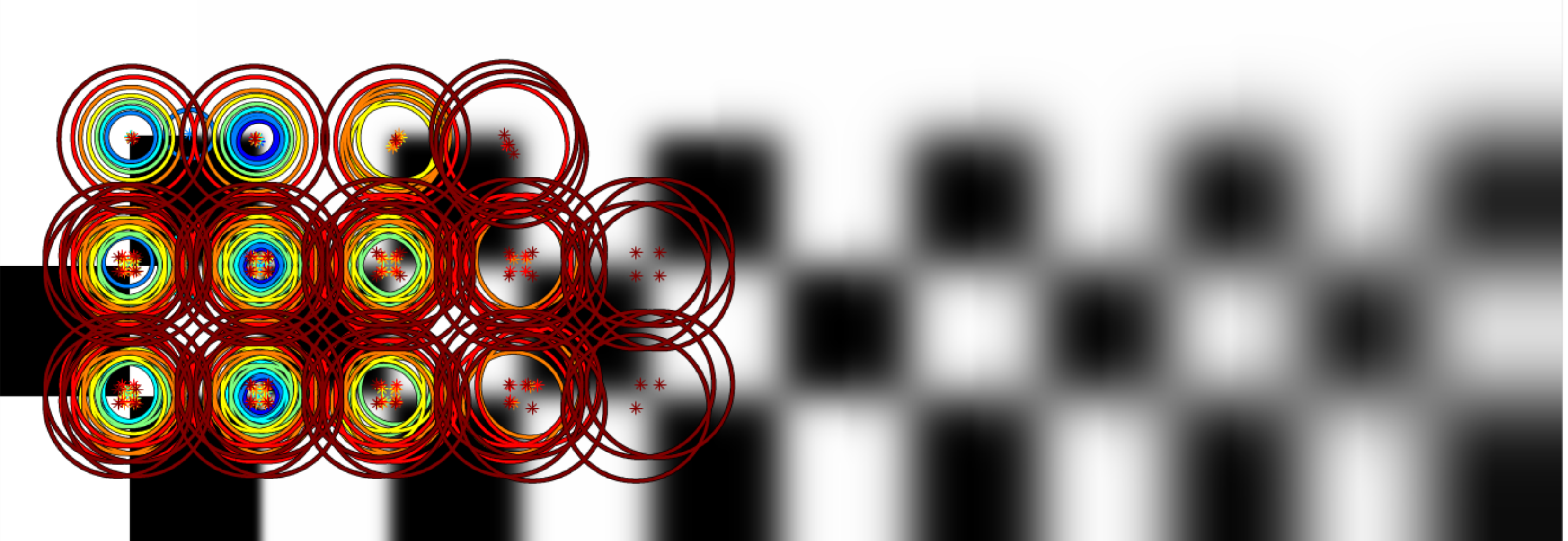}{ORB}

\caption{Detection on a progressively blurred chessboard pattern. Circle color reflects feature scale,  its size shows the extent of the description region.}
\label{fig.blur_chess}
\end{figure}

\subsection{Alternating pattern on the outer ring}

The second test considers the 16 pixels that approximate a circle of radius 3 around the central point. The outer ring is depicted in Fig.~\ref{fig:innerouter} in light blue. Let the pixels on the outer ring be denoted as $B = \{b_j\ |\ j=1 \ldots 16\}$. Each of the pixels in $B$ is labeled by one of three labels $\{d, s, l\}$. The labels are determined by the pixel intensity $I_{b_j}$, the central intensity at the saddle point $\rho$, and the method parameter offset $\varepsilon$ as follows
\begin{equation} \label{eqn:orlabel}
L_{b_j}=\left\{ \begin{array}{cc}
\textcolor{red}{\bullet}\ d, & I_{b_j}<\rho-\varepsilon \\
\textcolor{blue}{\bullet}\ s, & \rho-\varepsilon\leq I_{b_j}\leq\rho+\varepsilon \\
\textcolor{green}{\bullet}\ l, & I_{b_j}>\rho+\varepsilon 
\end{array}\right.
\end{equation}
The color of the dots in (\ref{eqn:orlabel}) corresponds to the color of the dots in the outer ring in Figs.~\ref{fig.saddle} and~\ref{fig:innerring}~(b).

The test is passed if the outer ring contains exactly two consecutive arcs of each label $l$ and $d$, the arcs are of length 2 to 8 pixels and are alternating -- the $l$ arcs are separated by $d$ arcs. To eliminate instability caused by $\rho$-crossing between $l$ and $d$ arcs, up to two pixels can be labeled $s$ at each boundary between $l$ and $d$ arcs. Labels $s$ are pixels with intensity in $\varepsilon$-neighborhood of $\rho$, where $\varepsilon$ is a parameter of the detector.

The test may seem complex, but in fact it is are regular grammar expression, which is equivalent to a finite state automaton and can be implemented very efficiently.

\begin{figure}
\newcommand{\ic}[2]{\setlength{\fboxsep}{0pt}%
\setlength{\fboxrule}{1pt}%
\parbox{\linewidth}{\centering \fbox{\includegraphics[width=0.8\linewidth]{images/#1}}\\#2}}

\ic{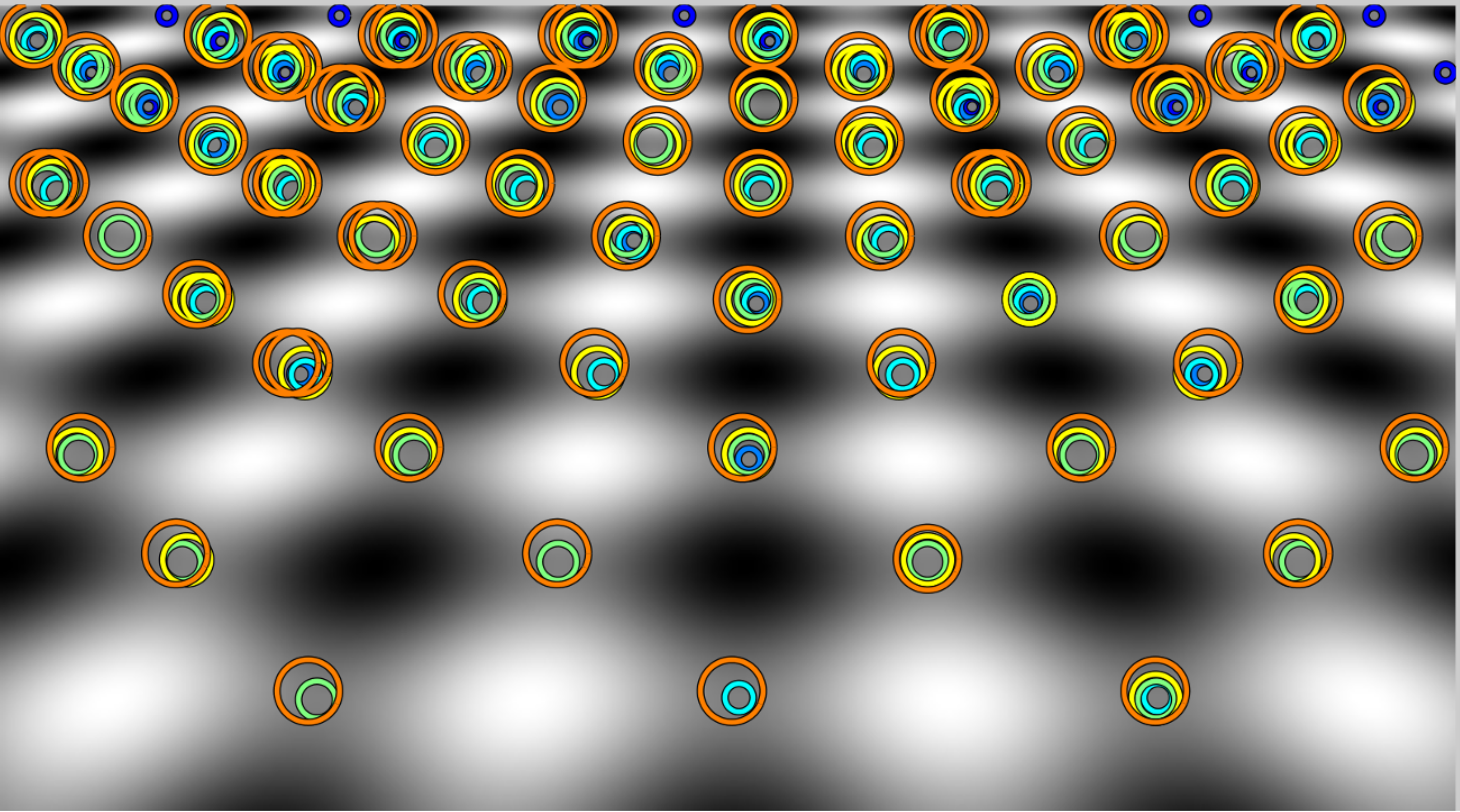}{(a) \dt}\\
\ic{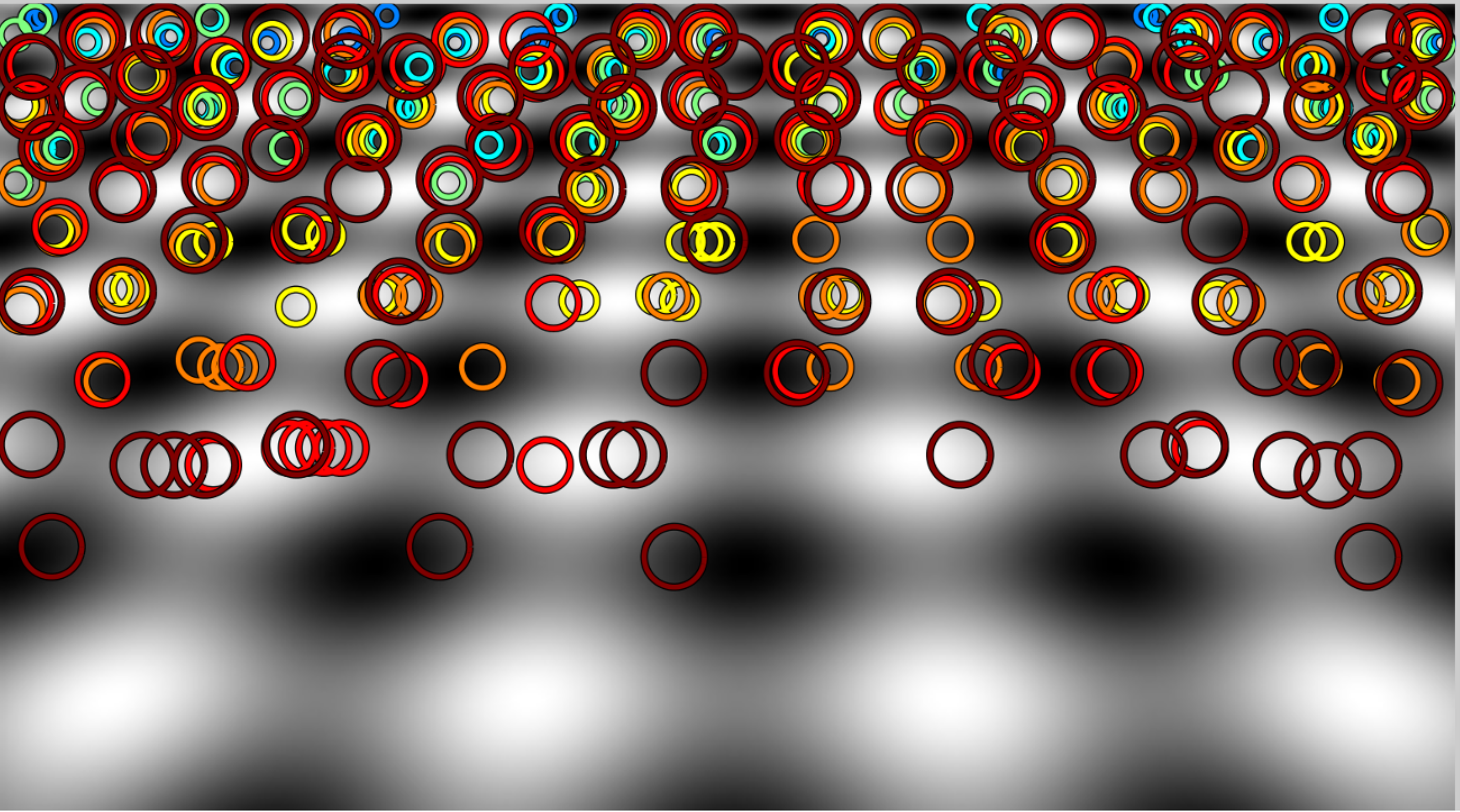}{(b) ORB}
\caption{Detection on a 2D sinusoidal pattern under a perspective transformation. \dt and ORB detections are shown as circles of the outer ring size.}
\label{fig.blur_blobs}
\end{figure}

% -------- coverage figure here -------------------------
\begin{figure*}
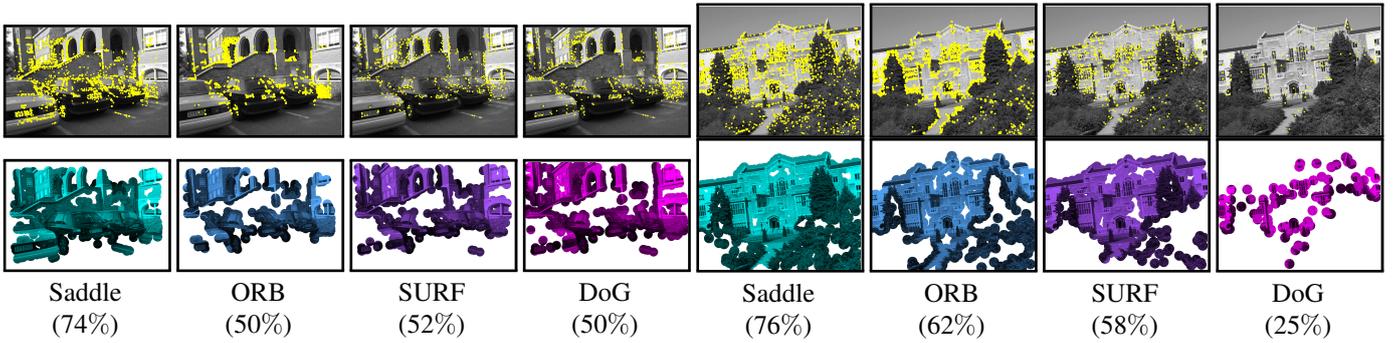


\newcommand{\igc}[5]{\parbox{.24\linewidth}{\centering \vspace{#5cm} \fbox{\includegraphics[width=0.99\linewidth]{images/ICenters_#1_pair_#2_dct_#3}}\\ \vspace{#5cm} \fbox{\includegraphics[width=0.99\linewidth]{images/IMask_#1_pair_#2_dct_#3}}\\#3\\(#4$\%$)}}
\newcommand{\ic}[7]{\parbox{.49\linewidth}{\igc{#1}{#2}{Saddle}{#3}{#7} \igc{#1}{#2}{ORB}{#4}{#7} \igc{#1}{#2}{SURF}{#5}{#7} \igc{#1}{#2}{DoG}{#6}{#7}}}

\setlength{\fboxsep}{0pt}%
\setlength{\fboxrule}{1pt}%
\ic{Light}{2}{74}{50}{52}{50}{0.28} \ic{Ubc}{4}{76}{62}{58}{25}{0}
% \ic{Bark}{1}{19}{12}{29}{32}{0}

\caption{Coverage by ground-truth validated feature matches  on selected image pairs from the Oxford dataset\cite{Mikolajczyk2005,Mikolajczyk2005a}. Yellow dots mark positions of the features (top). The covered area is computed as a union of circles with a 25 pixel radius centered on the matches(bottom).}
\label{fig.coverage}
\end{figure*}

\subsection{Post-processing}

Each point $\p$ that passed the alternating pattern test for both the inner and outer ring is assigned a response strength 
$$
R(\p) = \sum_{b_j \in B(\p)} |\rho_\p - b_j| \mbox{.}
$$
The value of the response strength is used in the non-maxima suppression step and to limit the number of responses if required.

\begin{SCfigure}
\centering
\caption{Positions of matched interest regions detected with \dt, ORB, SURF and DoG showing the detection complementarity.}
\setlength{\fboxsep}{0pt}%
\setlength{\fboxrule}{1pt}%
\fbox{\includegraphics[width=0.45\linewidth]{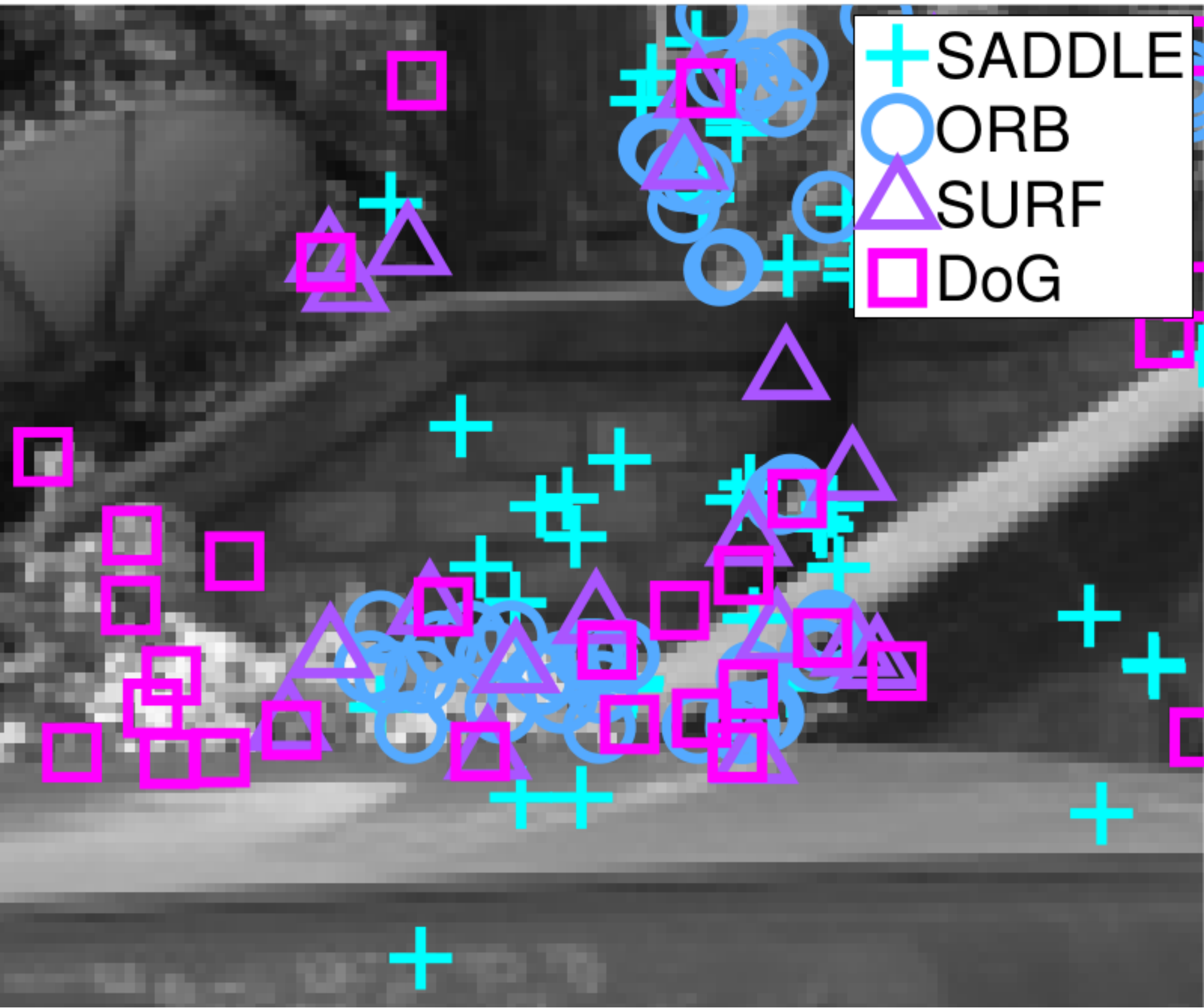}}
\label{fig:complement}
\end{SCfigure}

The non-maxima suppression is only performed within one level of the pyramid, features at different scales do not interact as the scale pyramid is relatively coarse. This is similar to non-maxima suppression of ORB. For the non-maxima suppression, a $3 \times 3$ neighborhood of point \p is considered.

As a final post-processing step, position refinement of points that passed the non-maxima suppression state takes place. A precise localization of the detected keypoint \p within the pyramid level is estimated with sub-pixel precision. The $x$ and $y$ coordinates of \p are computed as a weighted average of coordinates over a $3 \times 3$ neighborhood, where the weights are the response strengths $R$ of each pixel in the neighborhood. Response of pixels that do not pass the alternating pattern tests is set to 0. 

% ------------------------------------------------------------------------

\section{Experiments}
In this section, we experimentally evaluate the properties of the proposed \dt detector. The performance is compared with a number of commonly used feature detectors on standard evaluation benchmarks.

\subsection{Synthetic images}
\label{sec.synthetic_images}
Properties of the \dt and ORB, first are compared in two experiments on synthetically generated images.

First, features are detected on a chessboard pattern with progressively increasing blur, see Fig.~\ref{fig.blur_chess}. \dt point detection is expected in the central strips,
ORB detection on the corners on the right edge and potentially near the saddle points. \dt features are repeatedly detected at all blur levels and are well located at the intersection of the pattern edges. ORB features are missing at higher blur levels and their position is less stable.

A phenomenon common to corner feature points -- shifting from the corner for higher scales and blur levels is also visible.
Note that since the scaling factor between pyramid levels of \dt is 1.3 while for ORB it is 1.2,  \dt is run on a 6 level pyramid and ORB with 8 to achieve a similar range of scales.

Second, a standard synthetic test image introduced by Lindenberg and used in scale-space literature~\cite{Lindeberg91a} is used, see Fig.~\ref{fig.blur_blobs}. The \dt points are output at locations corresponding to saddle points across all scales in the perspectively distorted $f(\xi,\eta)  =\sin(\xi) \sin(\eta)$ pattern. Since there are no corners in the image, ORB detections are far from regular and are absent near the bottom edge. Fig.~\ref{fig:complement} shows the detector complementarity, i.e. \dt fires  on regions where other detectors have none detections.

%------------- coverage plots here -----------------
%% Graphs of Coverage vs image pair on KM dataset

\begin{figure*}
\newcommand{\ic}[1]{\parbox{.24\linewidth}{\centering #1\\ \vspace{0.1cm} \fbox{\includegraphics[width=0.99\linewidth]{images/IC_#1}}}}

\centering
\setlength{\fboxsep}{0pt}%
\setlength{\fboxrule}{1pt}%
\ic{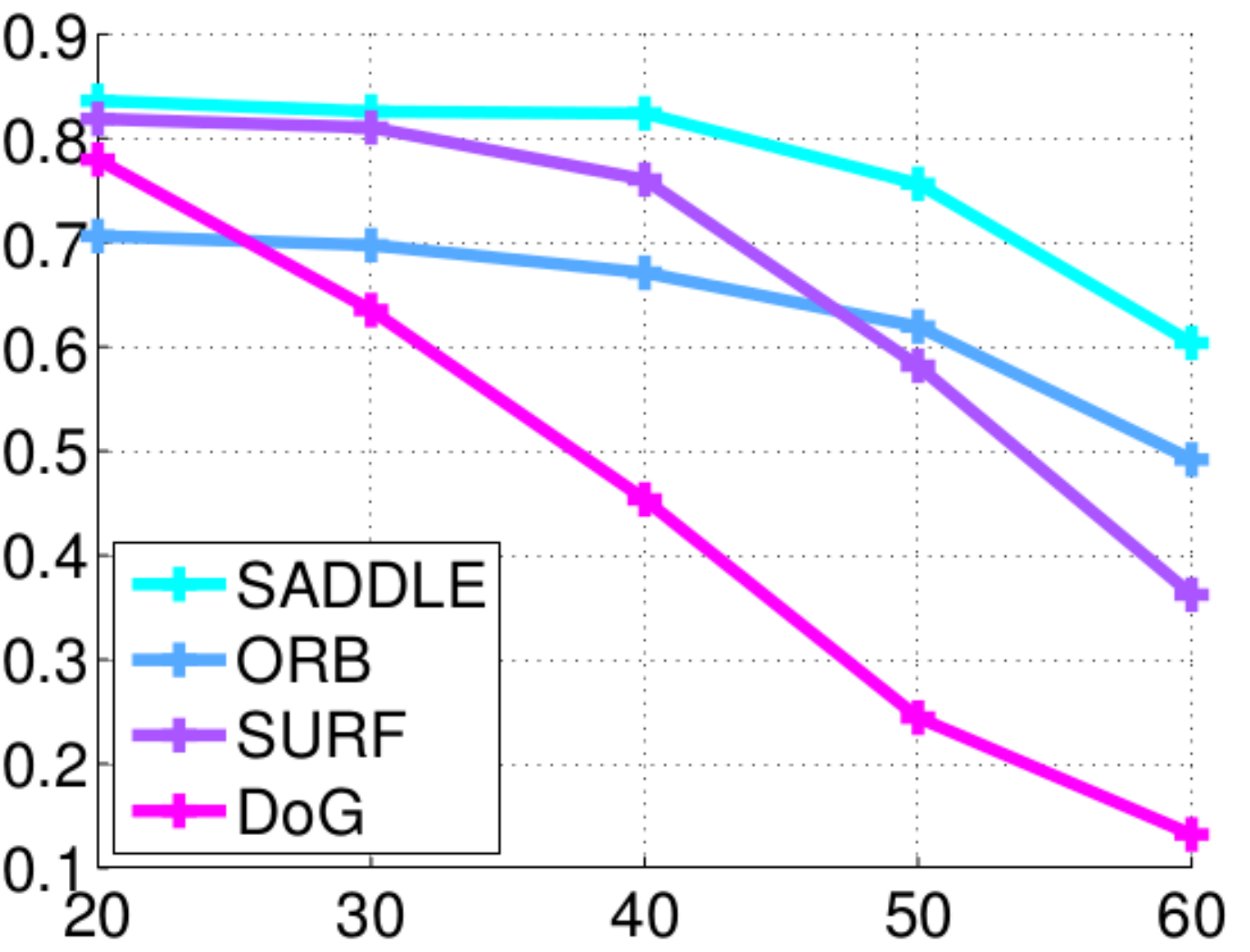} \hspace{0.5cm} \ic{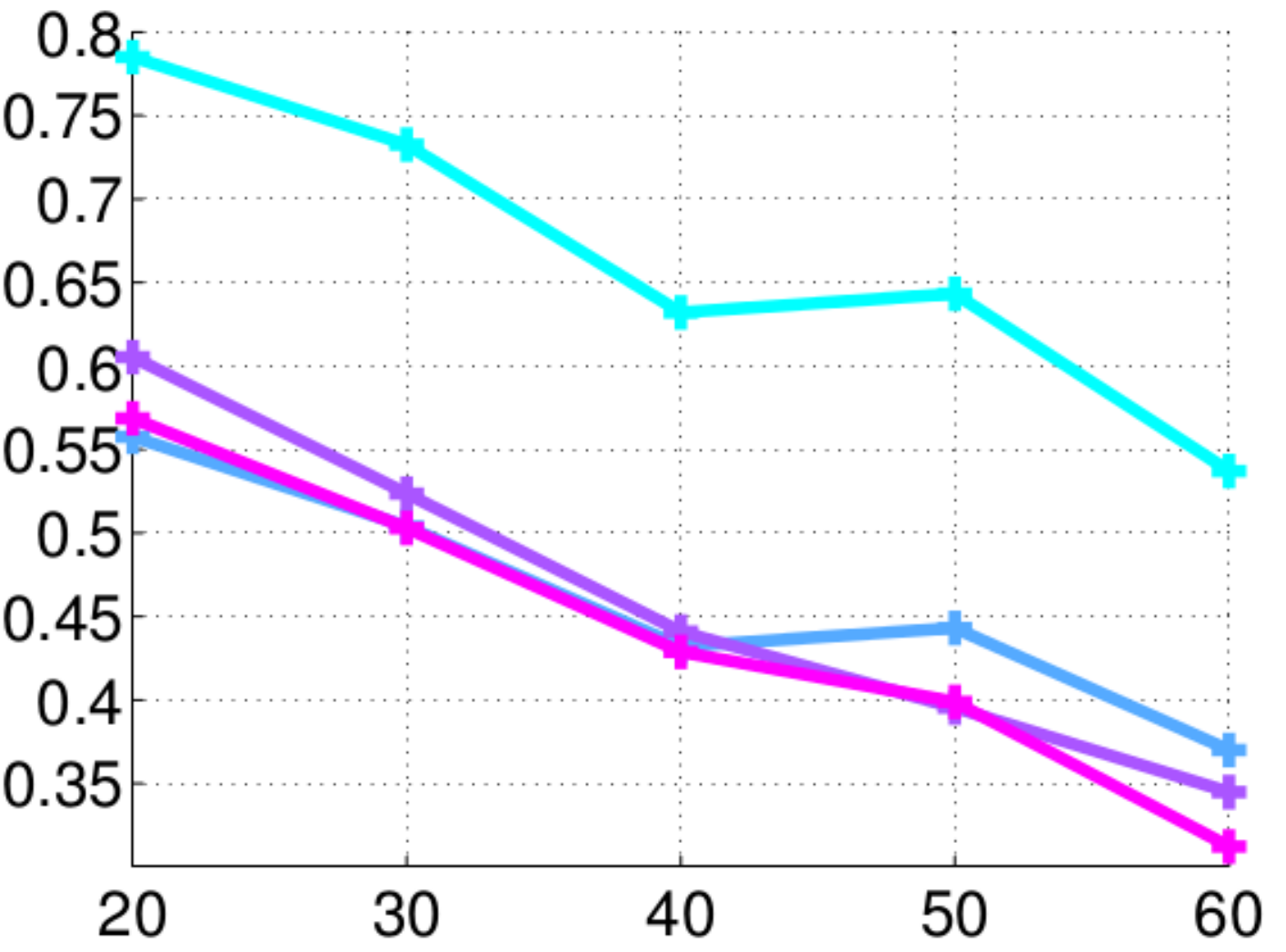} \hspace{0.5cm} \ic{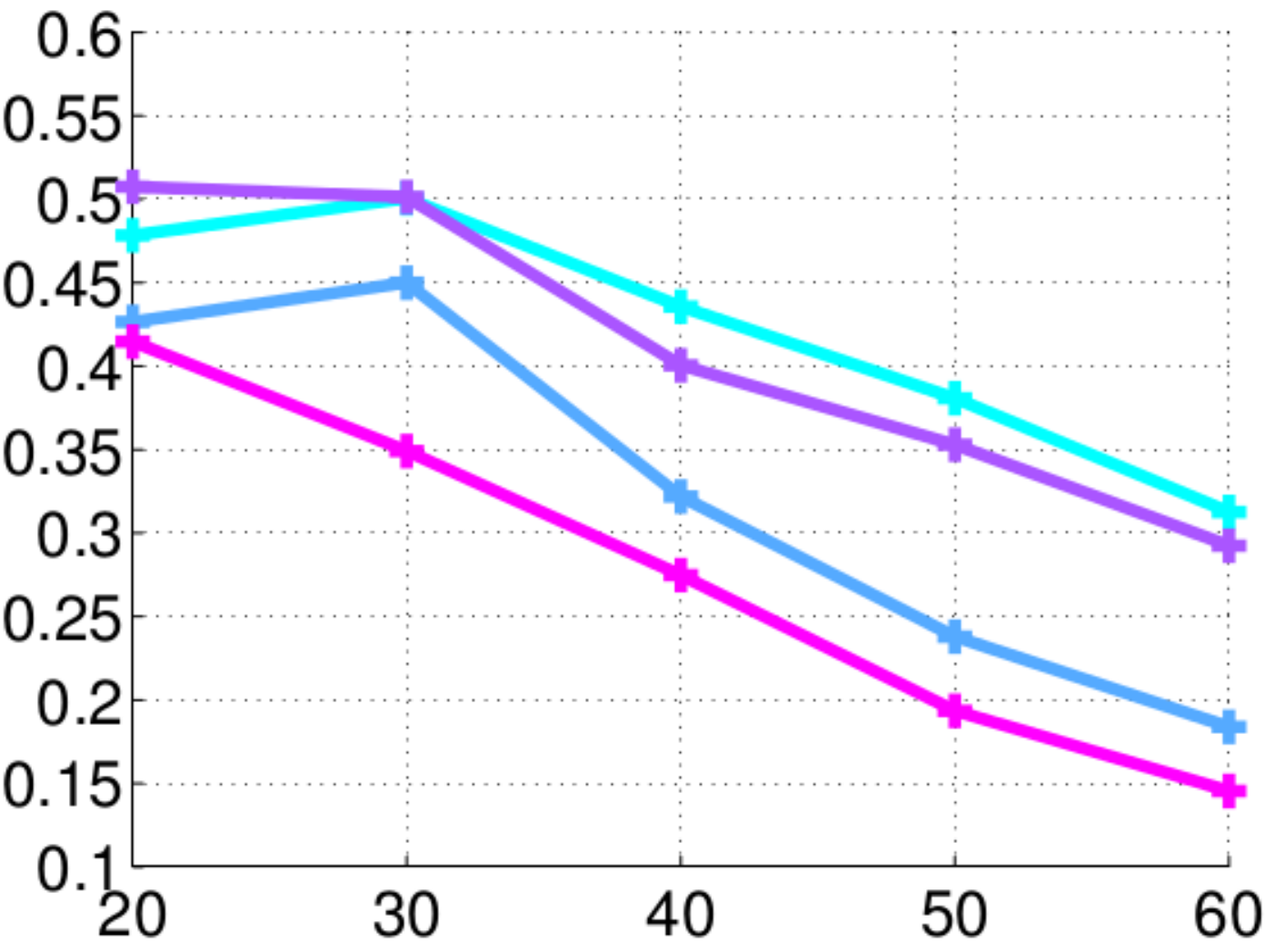}

\caption{Coverage by ground-truth validated feature matches on six image sets from the Oxford-Affine dataset~\cite{Mikolajczyk2005,Mikolajczyk2005a}. The x-axis shows the viewpoint angle and the y-axis shows the inlier coverage ratio in the reference image.}
\label{fig:coverageplots}
\end{figure*}

\subsection{Matching coverage}
In some task, such as structure from motion, good coverage of the image by matched point is crucial for the stability of the geometric models and consequently for the reliability of the 3D reconstruction~\cite{Irschara2009}. Note that the coverage is a~complementary criterion to the number of matched features, which is addressed in Section~\ref{sec:matches}. A high number of clustered matches may lead to poor geometry estimation and to incomplete 3D reconstruction.

To compare the coverage of different feature detectors, we adopt the measure proposed in \cite{Irschara2009}.
An image coverage mask is generated from matched features.
Every tentative correspondence geometrically consistent with the ground truth homography adds a disk of a fixed radius (of 25 pixels) into the mask at the location of the feature point.
The disk size does not change with the scale of the feature. The matching coverage is then measured as a fraction of the image covered by the coverage mask.

Extensive experiments show that the proposed \dt detector outperforms all other compared detectors: ORB, SURF and DoG.
Quantitative results are shown in Figure~\ref{fig:coverageplots}. The covered areas are shown in Figure~\ref{fig.coverage}.
The superior coverage of the \dt detector is visible on Fig.~\ref{fig:sad-orb-coverage}.
%%%%%%%%%%%%%%%%%%%%%%%%%%%%%%%%%%%%%%%%%%%%%%%%%%%%%%%%%%%%%
%%%% Graph of overlap in KM dataset 
\subsection{Accuracy}
\begin{figure}
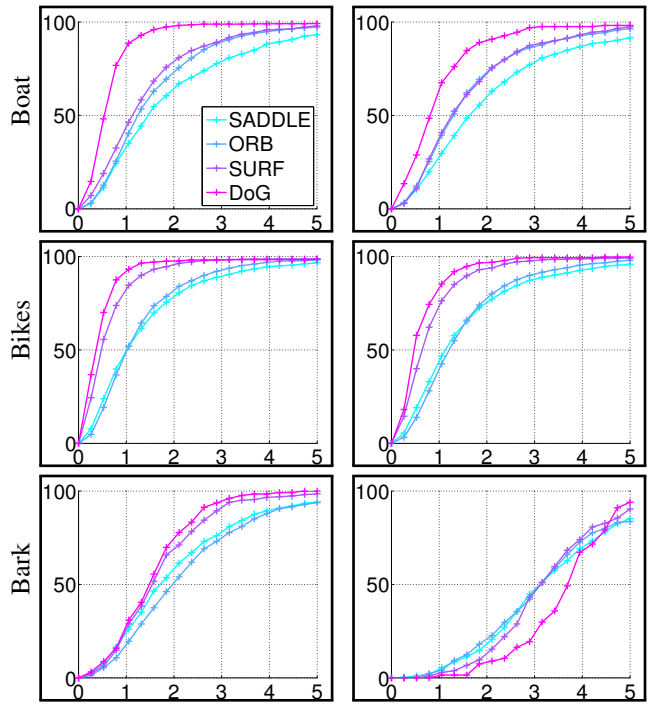


\newcommand{\igc}[2]{\rotatebox{90}{#1} \fbox{\parbox{3.8cm}{\includegraphics[width=.99\linewidth]{images/IR_#1_pair_#2}}}}
\newcommand{\ig}[2]{\fbox{\parbox{3.8cm}{\includegraphics[width=.99\linewidth]{images/IR_#1_pair_#2}}}}

\newcommand{\icc}[1]{\parbox{.5\linewidth}{\centering \igc{Boat}{#1}\\\vspace{0.1cm} \igc{Bikes}{#1}\\\vspace{0.1cm}  \igc{Bark}{#1}}}
\newcommand{\ic}[1]{\parbox{.18\linewidth}{\centering \ig{Boat}{#1}\\\vspace{0.1cm} \ig{Bikes}{#1}\\\vspace{0.1cm}  \ig{Bark}{#1}}}

\setlength{\fboxsep}{0pt}%
\setlength{\fboxrule}{1pt}%
\icc{1} \ic{2}

\caption{Inlier ratio (y-axis) curves on Oxford dataset \cite{Mikolajczyk2005}. The reprojection error is given in pixels (x-axis) in the reference image.}
\label{fig:inlier-ratio}
\end{figure}

The accuracy of \dt was assessed on the Oxford-Affine dataset. 
The cumulative distributions of reprojection errors with respect to the ground truth homography  are shown in Figure~\ref{fig:inlier-ratio}. \dt marginally outperforms ORB and DoG performance is superior in most cases.

%% Inlier ratio plotting on KM dataset

\subsection{Matching ability} \label{sec:matches}
In this section we follow the detector evaluation protocol from ~\cite{Mishkin2015WXBS}. We apply it to a restricted number of detectors -- those that are direct competitors of \dt: ORB~\cite{Rublee2011}, Hessian~\cite{Mikolajczyk2004} (extracting similar keypoints) and SURF~\cite{Bay2008} (also known as FastHessian).

We focus on getting a reliable answer to the match/no-match question for challenging image pairs. Performance is therefore measured by the number of successfully matched pairs, i.e. those with at least $15$ inliers found. The average number of inliers provides a finer indicator of the performance. 

\begin{figure*}
\setlength{\fboxsep}{0pt}%
\setlength{\fboxrule}{1pt}%
\centering
\parbox{.34\linewidth}{\centering \fbox{\includegraphics[width=0.99\linewidth]{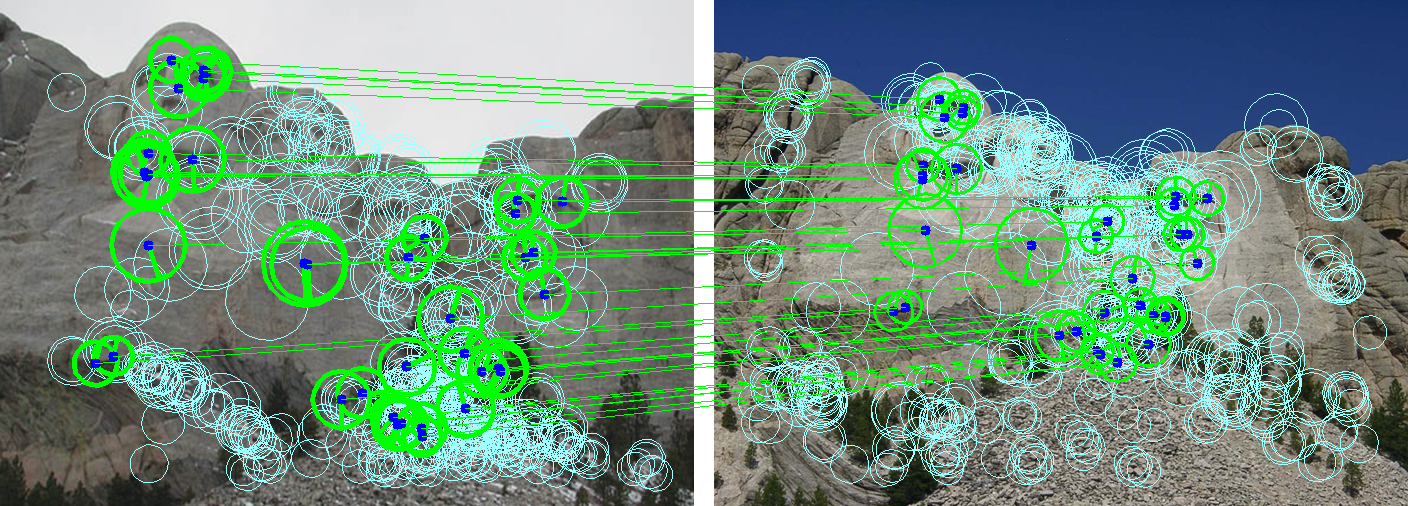}}\\\fbox{\includegraphics[width=0.99\linewidth]{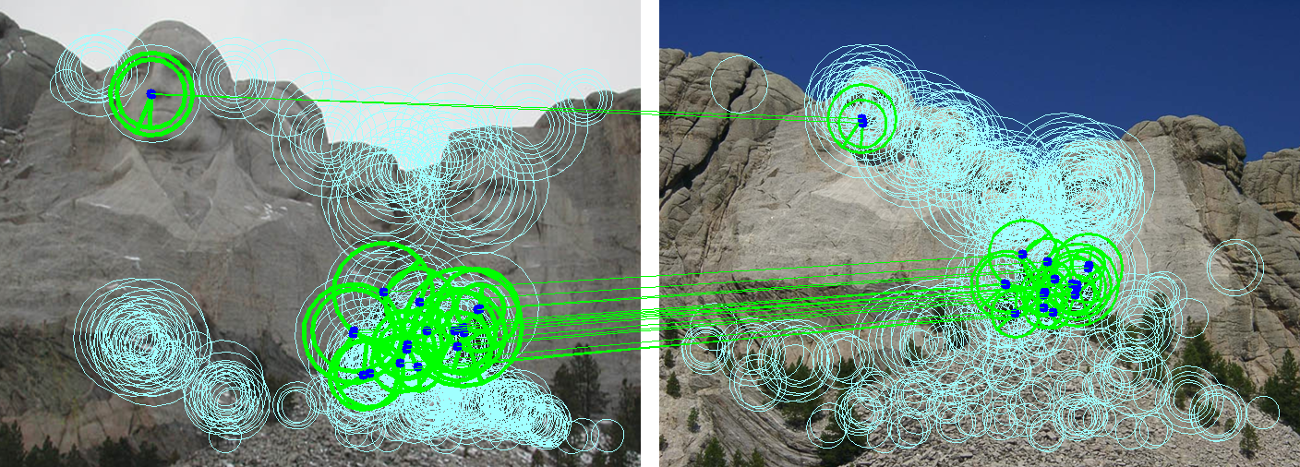}} \\ EF~\cite{Zitnick2011}. \dt: 37, ORB: 19.}
\parbox{.37\linewidth}{\centering\fbox{\includegraphics[width=0.99\linewidth]{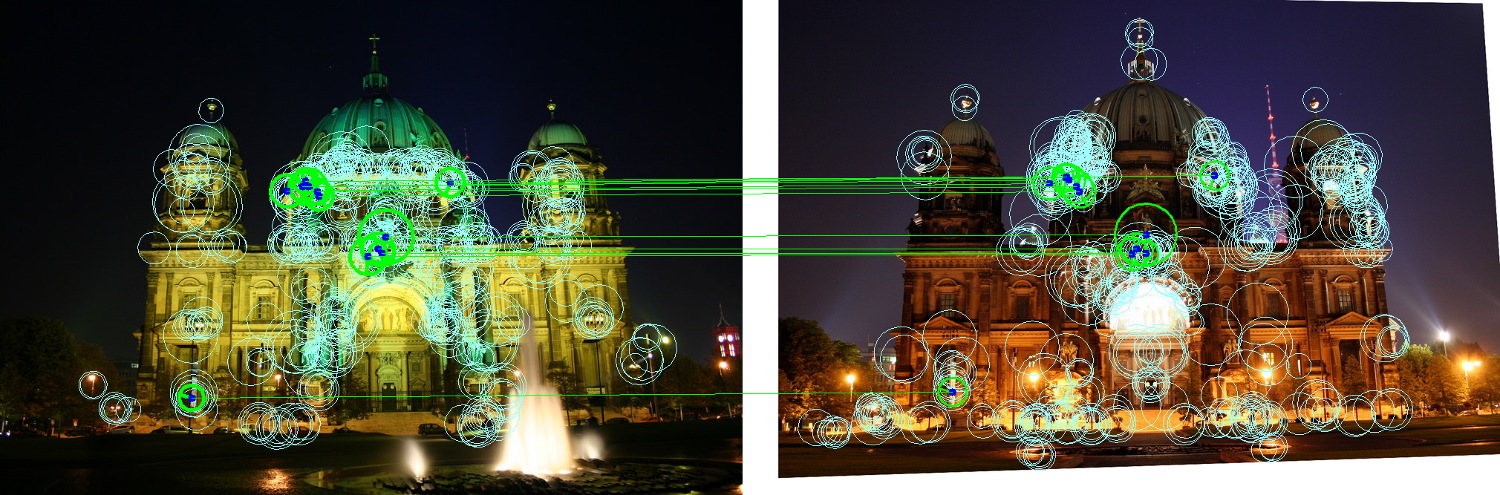}}\\\fbox{\includegraphics[width=0.99\linewidth]{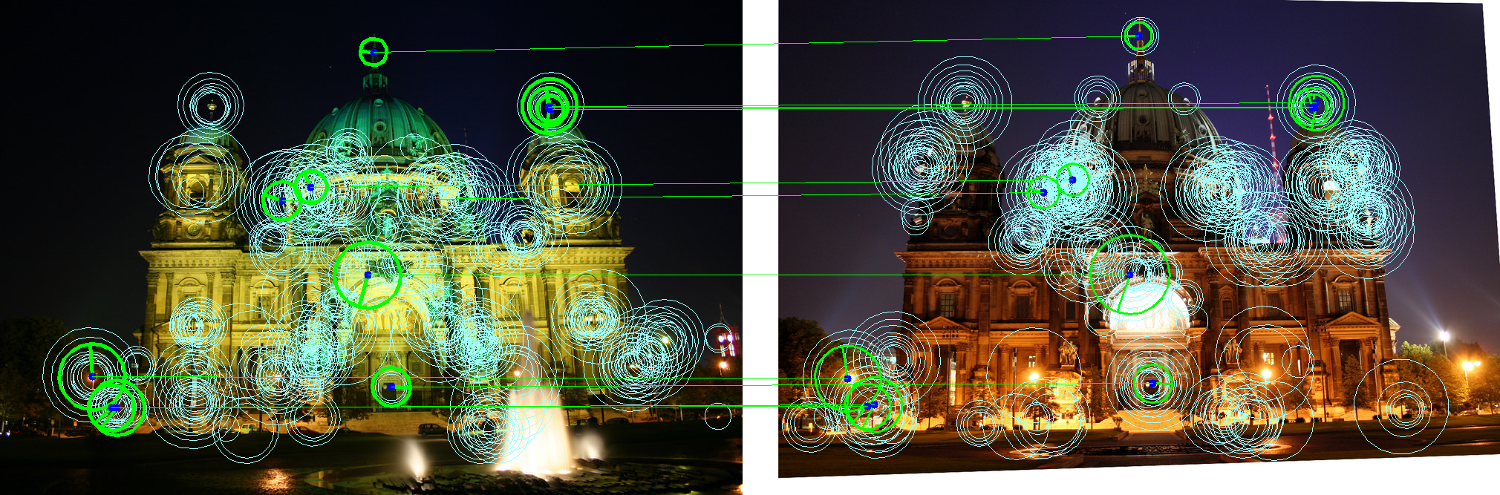}}\\
SymB~\cite{Hauagge2012}. \dt: 12, ORB: 11.}
\parbox{.26\linewidth}{\centering \fbox{\includegraphics[width=0.99\linewidth]{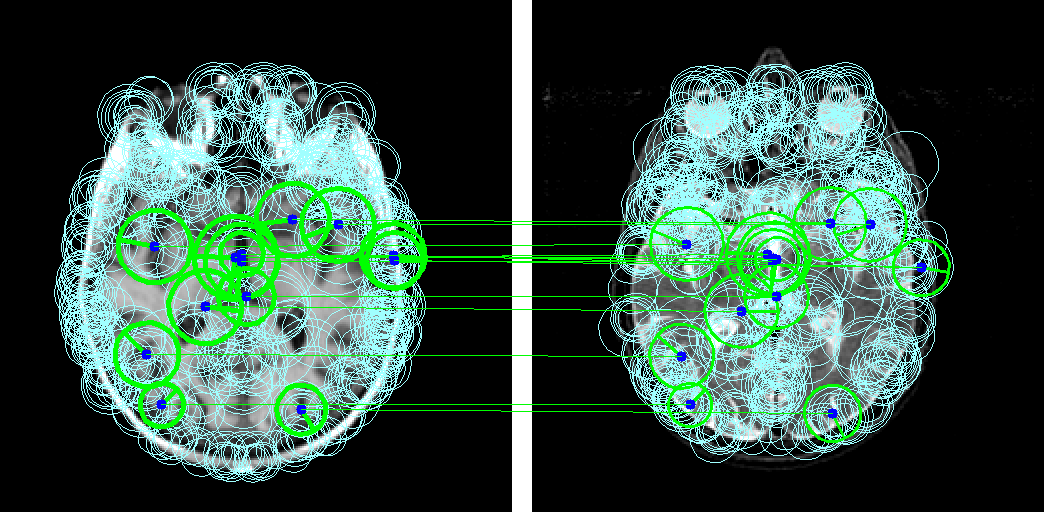}}\\\fbox{\includegraphics[width=0.99\linewidth]{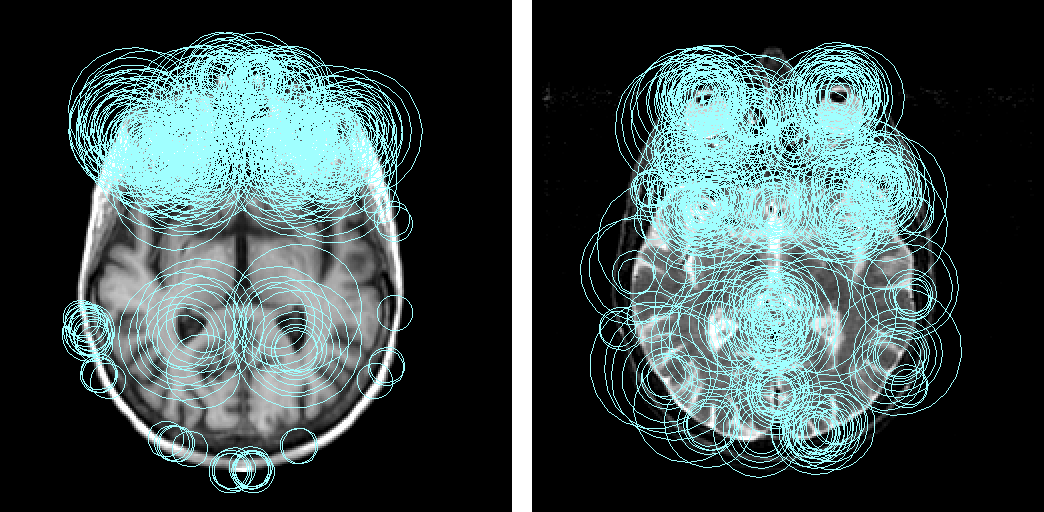}}\\
GDB~\cite{Kelman2007}, \dt: 13, ORB: 0.}%
%\parbox{.24\linewidth}{\centering \fbox{\includegraphics[width=0.99\linewidth]{images/adam-ORB.png}}\\\fbox{\includegraphics[width=0.99\linewidth]{images/adam-SADD.png}}}
\caption{Detected and matched keypoints for \dt (top) and ORB (bottom). The inliers count is given for both detectors for each image. Note that \dt points are spread more evenly making the homography estimation more stable. Images were selected from datasets listed in Table~\ref{tab:all-datasets}}
\label{fig:sad-orb-coverage}
\end{figure*}

The datasets used in this experiment are listed in Table~\ref{tab:all-datasets}. 
\begin{table}
\caption{Datasets used in evaluation}
\label{tab:all-datasets}
\scriptsize
\centering
\setlength{\tabcolsep}{.3em}
\begin{tabular}{llrl}
\hline
Short name& Proposed by& \#images&Nuisanse type\\
\hline
OxAff&Mikolajczyk~\etal\cite{Mikolajczyk2005},~\cite{Mikolajczyk2005a}, 2013&8x6&Geom., blur, illum.\\
EF&Zitnick and Ramnath~\etal\cite{Zitnick2011},2011&8x6&geom., blur, illum.\\
GDB&Kelman~\etal~\cite{Kelman2007}, 2007&22x2&illum., sensor\\
SymB&Hauagge and Snavely~\cite{Hauagge2012}, 2012&46x2&appearance\\
%LostInPast&Fernando~\etal\cite{Fernando2014}, 2014& 502&appearance, geom.\\
%EVD&Mishkin~\etal~\cite{Mishkin2015MODS}, 2015&15x2&Extreme geom.\\
\end{tabular}
\vspace{-1em}
\end{table}
Results are presented in two tables.
Table~\ref{tab:matching-sift} shows the results for a setup that focuses on matching speed and thus uses the fast BRIEF~\cite{Calonder2010} and FREAK~\cite{Alahi2012} descriptors (OpenCV implementation). Saddle works better with FREAK, while ORB results are much better with BRIEF. Saddle covers larger area and on broad class of images (e.g. see Figure~\ref{fig:sad-orb-coverage}), but needs different descriptor than BRIEF, possible optimized for description of saddle points. 

In an experiment \dt is run with a combination of RootSIFT~\cite{Arandjelovic2012} and HalfRootSIFT~\cite{Kelman2007} as descriptors (see Table~\ref{tab:matching-sift}).  This combination was claimed in recent benchmark~\cite{Mishkin2015WXBS} as best performing along broad range of datasets and it is suitable for evaluation of the matching potential of the feature detectors. With the powerful descriptors, \dt  clearly outperforms ORB. The MODS and WxBS are added as state-of art matchers in their original setup.
Most time is taken by description and matching.

Note that one could use both \dt and ORB detectors and benefit both from their speed and their complementarity (last rows in Table~\ref{tab:matching-sift}).

\subsection{Speed}
\begin{figure}
 \centering
\includegraphics[width=1\linewidth]{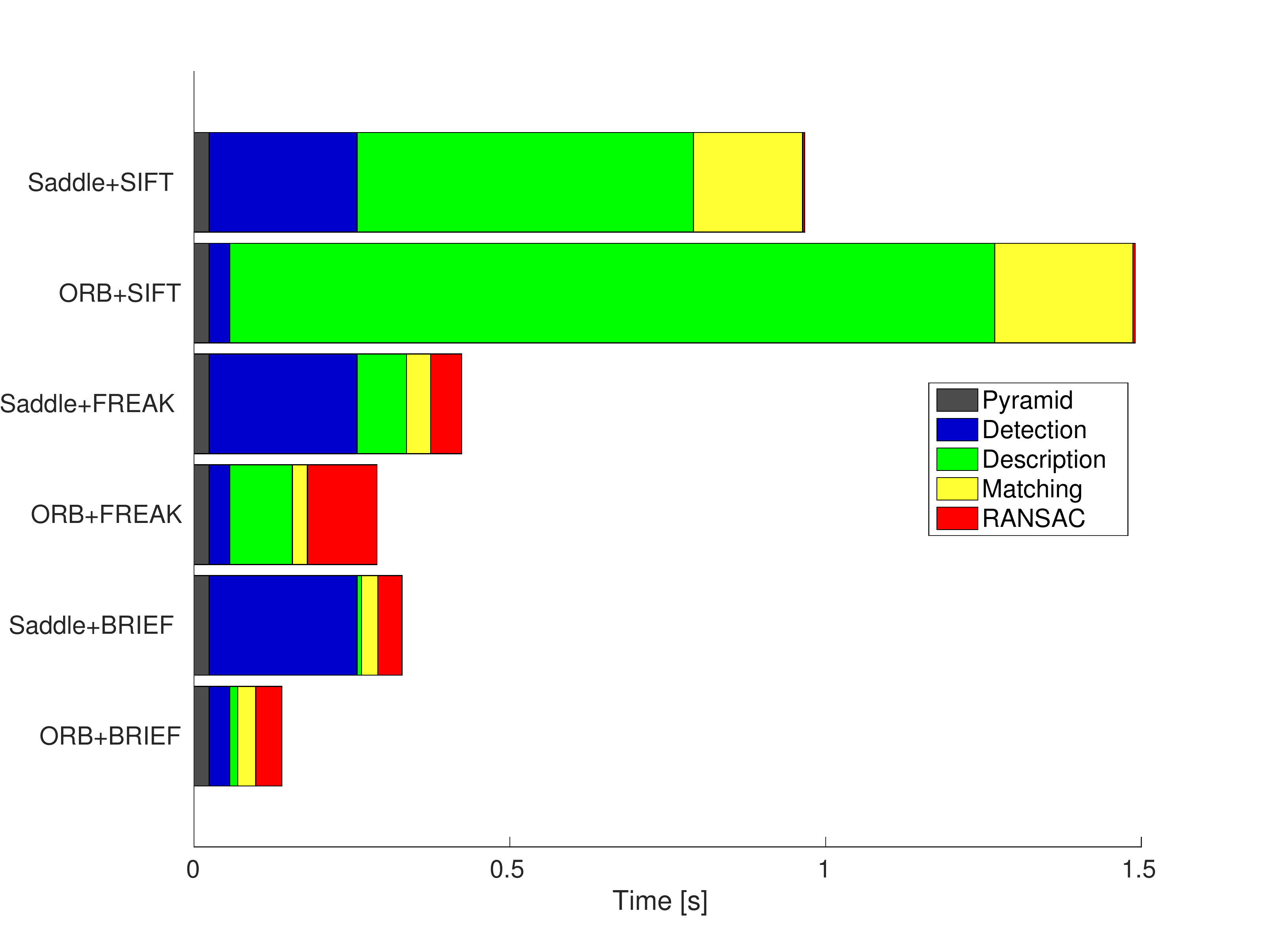}
\caption{Average run-time for ORB and Saddle on the Oxford-Affine dataset (average image size is $\approx$900x600 and average number of features is $\approx$1000).}
\label{fig:time-breakdown}
\end{figure}

The time breakdown for Saddle and ORB image matching on the Oxford-Affine
dataset is shown in Fig~\ref{fig:time-breakdown}. Saddle is about two times slower than ORB in the detection part. However, we have neither utilized SSE instructions in the Saddle tests. The results show that both Saddle and ORB are faster than the FREAK descriptor, but significantly slower than BRIEF. The slower RANSAC step for ORB with BRIEF is due to the lower inlier ratio.
 
%------------------------------------------------------------------------

% ------------------------------------------------------------------------

\section{Conclusion}
In this work we presented Saddle -- a novel similarity-covariant feature detector that responds to distinctive image regions at saddle points of the intensity function.

Experiments show that the \dt features are general, evenly spread ad appearing in high density in a range of images. 
The \dt detector is among the fastest proposed. 
In comparison with detector with similar speed, the \dt features show superior matching performance on number of challenging datasets.

\begin{table}
\centering
\small
\setlength{\tabcolsep}{.25em}
\vspace{0.5em}
\label{tab:matching-fast}
\caption{\dt evaluation with fast BRIEF and FREAK descriptors. The subcolumns are: the number of successfully matched image pairs (left),average running time (all stages: read image-detect-describe-match-RANSAC), average number of inliers in matched pairs (right). B stands for BRIEF, F -- for FREAK. The datasets are listed in Table~\ref{tab:all-datasets}. Darker cell background indicates better results.}
\begin{tabular}{|l|r|c|rrr|rrr|rrr|}
\hline
Alg.
&Sens
&Desc
& \multicolumn{3}{c|}{EF}
& \multicolumn{3}{c|}{OxAff}
& \multicolumn{3}{c|}{SymB}
\\
&&& \shortstack{33 \\\#}& \shortstack{time \\ $[s]$}& \shortstack{inl.  \\\#}
& \shortstack{40\\\#}& \shortstack{time \\ $[s]$}& \shortstack{inl.  \\\#}
& \shortstack{46\\\#}& \shortstack{time \\ $[s]$}& \shortstack{inl.  \\\#}
\\
\hline
Saddle&0.5K&B & \ccaEF{2} & 0.3 & 32 & \ccaoxford{22} & 0.3 & 47 & \ccasymbench{1} & 0.3 & 41\\
Saddle&0.5K&F & \ccaEF{11} & 0.3 & 29 & \ccaoxford{26} & 0.4 & 77 & \ccasymbench{8} & 0.4 & 32\\
ORB&0.5K&B & \ccaEF{16} & 0.1 & 28 & \ccaoxford{27} & 0.1 & 110 & \ccasymbench{14} & 0.2 & 30\\
ORB&0.5K&F & \ccaEF{0} & 0.1 & 0 & \ccaoxford{0} & 0.5 & 0 & \ccasymbench{0} & 0.2 & 0\\
\hline
Saddle&1K&B & \ccaEF{2} & 0.3 & 52 & \ccaoxford{25} & 0.3 & 92 & \ccasymbench{3} & 0.3 & 70\\
Saddle&1K&F & \ccaEF{14} & 0.4 & 47 & \ccaoxford{29} & 0.4 & 145 & \ccasymbench{11} & 0.4 & 59\\
ORB&1K&B & \ccaEF{18} & 0.2 & 52 & \ccaoxford{28} & 0.2 & 208 & \ccasymbench{21} & 0.2 & 51\\
ORB&1K&F & \ccaEF{0} & 0.2 & 0 & \ccaoxford{0} & 0.3 & 0 & \ccasymbench{0} & 0.2 & 0\\
\hline
Saddle+ORB&0.5K&B & \ccaEF{7} & 0.3 & 51 & \ccaoxford{27} & 0.4 & 140 & \ccasymbench{11} & 0.4 & 41\\
Saddle+ORB&0.5K&F & \ccaEF{19} & 0.4 & 42 & \ccaoxford{29} & 0.6 & 170 & \ccasymbench{14} & 0.5 & 51\\
\hline
Saddle+ORB&1K&B & \ccaEF{9} & 0.3 & 75 & \ccaoxford{28} & 0.4 & 269 & \ccasymbench{12} & 0.4 & 82\\
Saddle+ORB&1K&F & \ccaEF{18} & 0.4 & 87 & \ccaoxford{31} & 0.9 & 317 & \ccasymbench{21} & 0.5 & 85\\
\hline
\end{tabular}
%\vspace{1em}

\label{tab:matching-sift}.
\caption{\dt evaluation with a combination of RootSIFT and HalfRootSIFT descriptors. The subcolumns are the same as in Table~\ref{tab:matching-fast}. NMS stands for spatial non-maximum supression, indicating its application. In MODS-S, ORB was replaced by Saddle+FREAK, other parameters kept original.
Darker cell background indicates better results.}
\begin{tabular}{|l|r|c|c|rrr|rrr|rrr|rrr|rrr|rrr|}
\hline
Alg.&Sens&\begin{turn}{90}S\end{turn}&Desc
& \multicolumn{3}{c|}{EF}
& \multicolumn{3}{c|}{OxAff}
& \multicolumn{3}{c|}{SymB}
\\
&&\shortstack{\begin{turn}{90}M \end{turn}\\ \begin{turn}{90}N \end{turn}}&& \shortstack{33 \\\#}& \shortstack{time \\ $[s]$}& \shortstack{inl. \\\#}
& \shortstack{40\\\#}& \shortstack{time \\ $[s]$}& \shortstack{inl. \\\#}
& \shortstack{46\\\#}& \shortstack{time \\ $[s]$}& \shortstack{inl. \\\#}
\\
\hline
Saddle&0.5K&-&SIFT & \ccaEF{9} & 0.5 & 33 & \ccaoxford{30} & 0.6 & 73 & \ccasymbench{8} & 0.6 & 31\\
ORB&0.5K&- &SIFT& \ccaEF{13} & 0.7 & 35 & \ccaoxford{36} & 0.8 & 112 & \ccasymbench{12} & 0.8 & 46\\
\hline
Saddle&0.5K&+&SIFT& \ccaEF{18} & 0.5 & 46 & \ccaoxford{32} & 0.6 & 123 & \ccasymbench{15} & 0.6 & 49\\
ORB&0.5K&+ &SIFT& \ccaEF{12} & 0.7 & 37 & \ccaoxford{36} & 0.9 & 112 & \ccasymbench{13} & 0.8 & 44\\
\hline
Saddle&1K&-&SIFT& \ccaEF{18} & 0.9 & 49 & \ccaoxford{33} & 0.9 & 134 & \ccasymbench{17} & 0.9 & 48\\
ORB&1K&- &SIFT& \ccaEF{25} & 1.5 & 49 & \ccaoxford{37} & 1.5 & 222 & \ccasymbench{27} & 1.4 & 55\\
\hline
Saddle&1K&+&SIFT& \ccaEF{25} & 0.9 & 70 & \ccaoxford{34} & 0.9 & 249 & \ccasymbench{25} & 1.0 & 71\\
ORB&1K&+ &SIFT& \ccaEF{24} & 1.4 & 50 & \ccaoxford{37} & 1.5 & 222 & \ccasymbench{27} & 1.5 & 55\\
\hline
SURF&&&SIFT& \ccaEF{31} & 0.6 & 51 & \ccaoxford{37} & 1.3 & 483 & \ccasymbench{35} & 1.2 & 93\\
\hline
MODS&& &SIFT& \ccaEF{33} & 1.0 & 36 & \ccaoxford{40} & 0.5 & 163 & \ccasymbench{41} & 3.9 & 35\\
%MODS-S& &&SIFT& \ccaEF{33} & 4.6 & 154 & \ccaoxford{40} & 1.4 & 186 & \ccasymbench{33} & 10 & 149\\
MODS-S& &&SIFT&\ccaEF{33} & 1.8 & 34 & \ccaoxford{40} & 1.7 & 257 & \ccasymbench{43} & 4.6 & 40\\

WXBS&& &SIFT& \ccaEF{33} & 1.2 & 41 & \ccaoxford{40} & 1.2 & 149 & \ccasymbench{45} & 5.5 & 45\\
\hline

\end{tabular}
\\
\end{table}

\bibliographystyle{IEEEtran}
\bibliography{egbib}

\end{document}